\documentclass[10pt,twocolumn,letterpaper]{article}

\usepackage{iccv}
\usepackage{times}
\usepackage{epsfig}
\usepackage{graphicx}
\usepackage{amsmath}
\usepackage{amssymb}

\usepackage{booktabs}
\usepackage{multirow}

\usepackage{bm}
\usepackage{bbding}

\usepackage{color}

\usepackage{pifont}% http://ctan.org/pkg/pifont
\newcommand{\cmark}{\ding{51}}%
\newcommand{\xmark}{\ding{55}}%

\usepackage[pagebackref=true,breaklinks=true,letterpaper=true,colorlinks,bookmarks=false]{hyperref}

\iccvfinalcopy % *** Uncomment this line for the final submission

\def\iccvPaperID{10048} % *** Enter the ICCV Paper ID here

\ificcvfinal\fi

\begin{document}

%%%%%%%%% TITLE
% \title{Where to Look Also Matters: an Adaptive Focus Approach for Efficient Video Recognition}

\title{Adaptive Focus for Efficient Video Recognition}

\author{Yulin Wang\thanks{Equal contribution.}\ , Zhaoxi Chen$^*$\!, Haojun Jiang, Shiji Song, Yizeng Han, Gao Huang\thanks{Corresponding author.}\\
Department of Automation, BNRist, Tsinghua University, Beijing, China\\
% Beijing, China\\
{\tt\small \{wang-yl19, jhj20, hanyz18\}@mails.tsinghua.edu.cn, frozen.burning@gmail.com,}\\
{\tt\small \{shijis, gaohuang\}@tsinghua.edu.cn}
% \and
}

\maketitle
\ificcvfinal\thispagestyle{empty}\fi

%%%%%%%%% ABSTRACT
\begin{abstract}
   % Recent works have shown that the efficiency of video recognition can be improved by strategically selecting important frames to process. In this paper, we propose a spatial redundancy-based approach which not only outperforms these existing methods, but is compatible with them. Inspired by the fact that not all image regions in video frames are task-relevant, we reduce the computational cost by inferring the expensive high-capacity network only on a relatively small but informative patch of each frame, which is adaptively localized with reinforcement learning. Such a framework can be naturally augmented by further reducing the temporal redundancy, e.g., dynamically skipping less valuable frames. Extensive empirical results on five benchmarks (i.e., ActivityNet, FCVID, Mini-Kinetics, Something-Something V1\&V2) demonstrate that our method is considerably more efficient than the competitive baselines as an alternative, while it can be built on the top of state-of-the-art light-weighted networks (e.g., TSM) to effectively improve their efficiency.
   In this paper, we explore the spatial redundancy in video recognition with the aim to improve the computational efficiency. It is observed that the most informative region in each frame of a video is usually a small image patch, which shifts smoothly across frames. Therefore, we model the patch localization problem as a sequential decision task, and propose a reinforcement learning based approach for efficient spatially adaptive video recognition (AdaFocus). In specific, a light-weighted ConvNet is first adopted to quickly process the full video sequence, whose features are used by a recurrent policy network to localize the most task-relevant regions. Then the selected patches are inferred by a high-capacity network for the final prediction. During offline inference, once the informative patch sequence has been generated, the bulk of computation can be done in parallel, and is efficient on modern GPU devices. In addition, we demonstrate that the proposed method can be easily extended by further considering the temporal redundancy, e.g., dynamically skipping less valuable frames. Extensive experiments on five benchmark datasets, i.e., ActivityNet, FCVID, Mini-Kinetics, Something-Something V1\&V2, demonstrate that our method is significantly more efficient than the competitive baselines. Code is available at \url{https://github.com/blackfeather-wang/AdaFocus}.
   % The code will be released upon the acceptance of the paper.
   % It can also be built on top of state-of-the-art light-weighted networks (e.g., TSM) to further improve the efficiency.
   % The full model can be trained conveniently in an end-to-end manner. 
   % improve their efficiency.
   % more efficient than 
   % to improve their efficiency
   % without sacrificing accuracy.
   % state-of-the-art
   % skipping uninformative video frames.
   % we adaptively identifies the most task-relevant regions of each frame with reinforcement learning, and allocate the majority of computations to these small but more valuable inputs. Such a framework
   % This paper proposes a novel adaptive focus (AdaFocus) framework for efficient video recognition. While existing methods mainly focus on selecting important frames to process, we demonstrate by AdaFocus that high computational efficiency can also be achieved by reducing spatial redundancy. In specific, our method merely process 
   % adaptively identifies the task-relevant regions of each frame with reinforcement learning, and merely perform 
   % allocating the majority of computations to these 
   % which leverages the cheap global information exacted by a light-weighted 
\end{abstract}

%%%%%%%%% BODY TEXT

% \vspace{-0.2in}
\section{Introduction}

The explosive growth of online videos (e.g., on YouTube or TikTok) has fueled the demands for automatically recognizing human actions, events, or other contents within them, which benefits applications like recommendation \cite{davidson2010youtube, deldjoo2016content, gao2017unified}, surveillance \cite{collins2000system, chen2019distributed} and content-based searching \cite{ikizler2007searching}. In the past few years, remarkable success in accurate video recognition has been achieved by leveraging deep networks \cite{feichtenhofer2019slowfast, zhu2017deep, feichtenhofer2016convolutional, carreira2017quo, tran2015learning, hara2018can}. However, the impressive performance of these models usually comes at high computational costs. In real-world scenarios, computation directly translates into power consumption, carbon emission and practical latency, which should be minimized under both economic and safety considerations.

\begin{figure}[t]
    % \vskip -0.1in
    \begin{center}
    \centerline{\includegraphics[width=0.935\columnwidth]{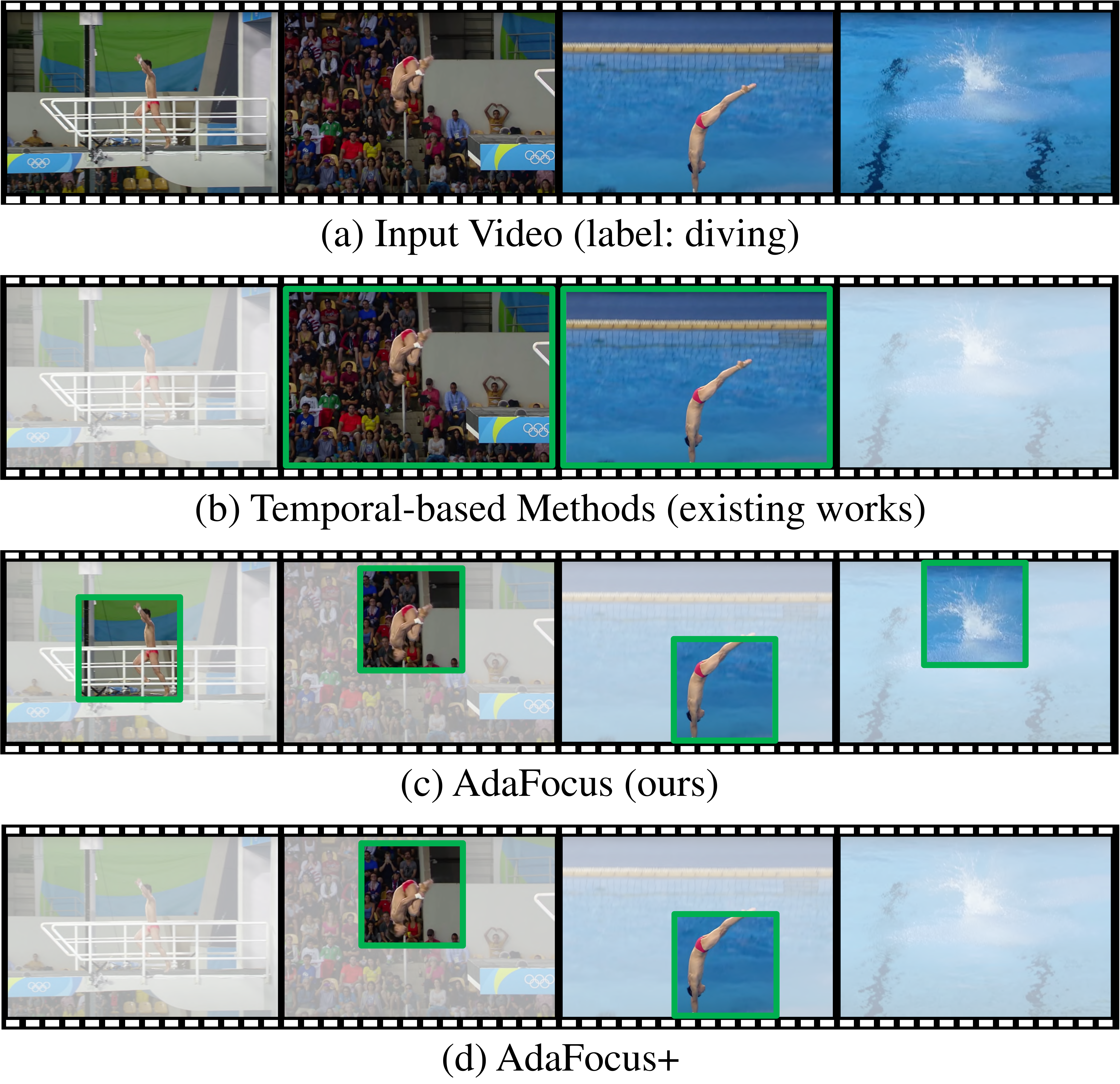}}
    % \vskip -0.04in
    \caption{\textbf{Comparisons of temporal-based methods and our proposed AdaFocus approach.} Most existing works reduce computational costs by selecting a few informative frames to process, while AdaFocus aims to perform efficient inference by attending to the task-relevant patch of each frame. Importantly, our method is compatible with temporal-based techniques as it can be improved by skipping less important frames (AdaFocus+). \label{fig:fig1}
    }
    \end{center}
    \vspace{-6ex}
    % \vskip -0.3in
\end{figure}

To address this issue, a number of recent works propose to reduce the inherent \emph{temporal redundancy} in video recognition \cite{li20202d, meng2020ar, wu2019liteeval, wu2019multi, gao2020listen, wu2019adaframe, korbar2019scsampler}. As shown in Figure~\ref{fig:fig1} (b), it is efficient to focus on the most task-relevant video frames, and allocate the majority of computation to them rather than all frames. However, another important source of redundant computation in image-based data, namely \emph{spatial redundancy}, has rarely been explored in the context of efficient video recognition. In fact, it has been shown in 2D-image classification that convolutional networks (CNNs) are able to produce correct predictions with only a few discriminative regions of the whole image \cite{wang2020glance, xie2020spatially, han2021dynamic, mnih2014recurrent, fu2017look, chu2019spot}. By performing inference on these relatively small regions, one can dramatically reduce the computational cost of CNNs (e.g., processing a 96x96 patch requires $\sim$18\% computation of inferring a 224x224 image).

In this paper, we are interested in whether this \emph{spatial redundancy} can be effectively leveraged to facilitate efficient video recognition.  We develop a novel adaptive focus (AdaFocus) approach to dynamically localize and attend to the task-relevant regions of each frame. In specific, our method first takes a quick glance at each frame with a light-weighted CNN to obtain cheap and coarse global information. Then we train a recurrent policy network on its basis to select the most valuable region for recognition. This procedure leverages the reinforcement learning algorithm due to the non-differentiability of localizing task-relevant regions. Finally, we activate a high-capacity deep CNN to process only the selected regions. Since the proposed regions are usually small patches with a reduced size, considerable computational costs can be saved. An illustration of AdaFocus is shown in Figure~\ref{fig:fig1} (c). Our method allocates computation unevenly across the spatial dimension of video frames according to the contributions to the recognition task, leading to a significant improvement in efficiency with preserved accuracy.

% , i.e., can we effectively prevent redundant computation without sacrificing accuracy via adaptively localizing and attending to the task-relevant regions of each frame. To this end, we develop an novel adaptive focus (AdaFocus) approach. 

The vanilla AdaFocus framework does not model temporal redundancy, i.e., all frames are processed with identical computation, while the only difference lies in the locations of the selected regions. We show that our method is compatible with existing temporal-based techniques, and can be extended via reducing the computation spent on uninformative frames, as presented in Figure~\ref{fig:fig1} (d). This is achieved by introducing an additional policy network that determines whether to skip some less valuable frames. This algorithm is referred to as AdaFocus+.

We evaluate the effectiveness of AdaFocus on five video recognition benchmarks (i.e., ActivityNet, FCVID, Mini-Kinetics, Something-Something V1\&V2). Experimental results show that AdaFocus by itself consistently outperforms all the baselines by large margins, while AdaFocus+ further improves the efficiency. For instance, AdaFocus+ has 2-3x less FLOPs\footnote{In this paper, FLOPs refers to the number of multiply-add operations.} than the recently proposed AR-Net \cite{meng2020ar} when achieving the same accuracy. We also demonstrate that our method can be deployed on top of the state-of-the-art networks (e.g., TSM \cite{lin2019tsm}) and effectively improve their computational efficiency.

\begin{figure*}[t]
    % \vskip -0.1in
    \begin{center}
    \centerline{\includegraphics[width=1.8\columnwidth]{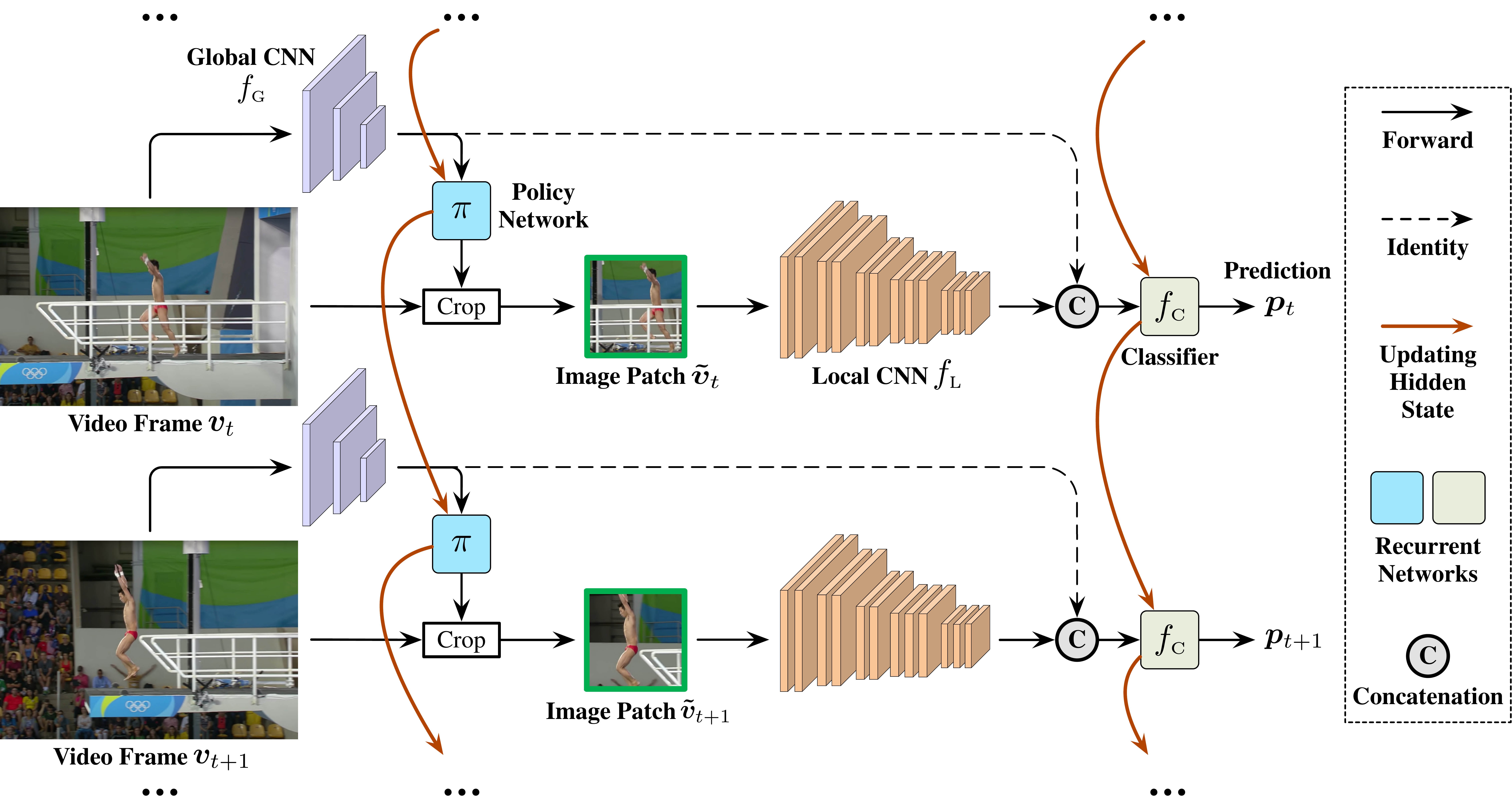}}
    % \vskip -0.025in
    \caption{\textbf{Overview of AdaFocus}. It first takes a quick glance at each frame $\bm{v}_t$ using a light-weighted global CNN $f_{\textnormal{G}}$. Then a recurrent policy network $\pi$ is built on top of $f_{\textnormal{G}}$ to select the most important image region $\tilde{\bm{v}}_t$ in terms of recognition. A high-capacity local CNN $f_{\textnormal{L}}$ is adopted to extract features from $\tilde{\bm{v}}_t$. Finally, a recurrent classifier aggregates the features across frames to obtain the prediction $\bm{p}_t$. \label{fig:overview}
    }
    \end{center}
    \vspace{-4.5ex}
    % \vskip -0.3in
\end{figure*}

\section{Related Works}

\textbf{Video recognition.} Significant progress has been made in video recognition with the adoption of convolutional neural networks (CNNs). One prevalent approach is constructing 3D-CNNs to model the temporal and spatial information jointly, e.g., C3D \cite{tran2015learning}, I3D \cite{carreira2017quo} and ResNet3D \cite{hara2018can}. The other line of works first extracts frame-level features, and aggregates the features at different temporal locations via temporal averaging \cite{wang2016temporal}, long short-term memory (LSTM) networks \cite{donahue2015long}, channel shifting \cite{lin2019tsm}, etc.

Despite the success achieved by the aforementioned works, the expensive computational cost of CNNs, especially 3D-CNNs, usually limits their applicability. Recent research efforts have been made towards improving the efficiency of video recognition, via designing light-weighted architectures \cite{tran2018closer, xie2018rethinking, pan2021va, tran2019video, zolfaghari2018eco, lin2019tsm} or perform dynamic computation on a per-video basis \cite{yeung2016end,wu2019liteeval,korbar2019scsampler,zhu2020a3d,li20202d,meng2021adafuse}. Our approach shares a similar idea as the latter on reducing the intrinsic redundancy in video data, while with a special focus on spatial redundancy.

% while can be built upon existing efficient network architectures, as we will show in this paper.

% Despite the success achieved by the aforementioned work, the expensive computational cost limits their deployment on resource-constrained platforms. Recent research efforts have been made towards improving the efficiency of video recognition. Specifically, existing arts either design lightweight architectures \cite{tran2018closer, xie2018rethinking, pan2021va, tran2019video, zolfaghari2018eco} or perform dynamic computation on a per-video basis \cite{yeung2016end,wu2019liteeval,korbar2019scsampler,zhu2020a3d,li20202d,meng2021adafuse}. Our approach shares the same goal as the latter on reducing the intrinsic redundancy in data, and can be built upon existing efficient network architectures, as we will show in our paper.

\textbf{Reducing temporal redundancy} is a popular solution to efficient video recognition, based on the intuition that not all frames contribute equally to the final prediction. In particular, a model could dynamically allocate no/little computation to some less informative or highly correlated frames \cite{han2021dynamic}. This idea has been proven effective by many implementations, including (1) {early stopping}, i.e., terminating the computation before ``watching'' the full sequence \cite{fan2018watching,wu2020dynamic}; (2) {conditional computing}, e.g., LiteEval \cite{wu2019liteeval} adaptively selects an LSTM model with appropriate size at each time step in a recurrent recognition procedure; adaptive resolution network (AR-Net) \cite{meng2020ar} processes different frames with adaptive resolutions to save unnecessary computation on less important frames; and (3) {frame/clip sampling}, i.e. dynamically deciding which frames should be skipped without performing any computation \cite{wu2019multi,gao2020listen,korbar2019scsampler,wu2019adaframe}. The proposed AdaFocus method differentiates itself from these approaches in that we focus on reducing the \emph{spatial redundancy}, namely allocating the major computation to task-relevant regions of the frames. In addition, our method is compatible with them as it can be improved by further reducing temporal redundancy.

% Compared to these approaches, our method focuses on the \emph{spatial} dimension
% reduces redundant computation from both \emph{spatial} and temporal dimensions and further improves the overall efficiency.

\textbf{Reducing spatial redundancy.} 
It has been observed that considerable spatial redundancy exists in the process of extracting deep features from image-based data \cite{han2021dynamic, yang2020resolution}. For example, in 2D-image classification, a number of recent works successfully improve the efficiency of CNNs by attending to some task-relevant or more informative parts of images \cite{xie2020spatially, wang2020glance, figurnov2017spatially}. In the context of video recognition, existing attention-based methods have demonstrated that different image regions of video frames do not contribute equivalently to the prediction \cite{meng2019interpretable}. However, to our best knowledge, the \emph{spatial redundancy} has not been exploited to improve the efficiency of video recognition.

% In computer vision, extensive efforts have been made to reduce spatial redundancy in CNNs for image data \cite{han2021dynamic}. 

% The most promising direction is developing dynamic networks which conduct adaptive computation on pixels \cite{xie2020spatially}, patches \cite{wang2020glance} or features with different resolutions \cite{yang2020resolution}.

% Among them, the recent glance-and-focus network (GFNet) \cite{wang2020glance} learns to perform image classification by recurrently focusing on a selected patch, achieving a significant improvement for the trade-off between accuracy and efficiency. 

% Apart from the commonly noticed temporal dimension, few researches have exploited the spatial redundancy for efficient video recognition. 

% % The recent adaptive resolution network (AR-Net) \cite{meng2020ar} represents different frames with adaptive resolutions to avoid the unnecessary computation induced by high-resolution feature maps for the "unimportant" frames. 

% In contrast to AR-Net that still treat different locations in a frame with uniform operations, our approach is able to allocate the computation to the most salient spatial locations, enjoying notably flexibility when processing each frame in a video.

\section{Method}
\label{sec:method}

Different from most existing works that facilitate efficient video recognition by leveraging the \emph{temporal redundancy}, we seek to save the computation spent on the task-irrelevant regions of video frames, and thus improve the efficiency by reducing the \emph{spatial redundancy}. To this end, we propose an adaptive focus (AdaFocus) framework to adaptively identify and attend to the most informative regions of each frame, such that the computational cost can be significantly reduced without sacrificing accuracy.

% In this section, we introduce the details of our method. We describe its components and the correspond training algorithm in Section \ref{sec:arch} and Section \ref{sec:training}, respectively. In addition, we show in Section \ref{sec:discussion} that AdaFocus can be naturally improved by further reducing temporal redundancy (e.g., skipping uninformative frames) or applying the dynamic inference technique, namely allocating computation unevenly across easy and hard samples at test time.

In this section, we first describe its components and the correspond training algorithm in Section \ref{sec:arch} and Section \ref{sec:training}, respectively. Then we show in Section \ref{sec:compatibility} that AdaFocus can be improved by further considering temporal redundancy (e.g., skipping uninformative frames).

% or applying the dynamic inference technique, namely allocating computation unevenly across easy and hard samples at test time.

\subsection{Network Architecture}
\label{sec:arch}

\textbf{Overview.}
We first give an overview of AdaFocus (Figure \ref{fig:overview}). Consider the online video recognition scenario, where a stream of frames come in sequentially while a prediction may be retrieved after processing any number of frames. At each time step, AdaFocus first takes a quick glance at the full frame with a light-weighted CNN $f_{\textnormal{G}}$, obtaining cheap and coarse global features. Then the features are fed into a recurrent policy network $\pi$ to aggregate the information across frames and accordingly determine the location of an image patch to be focused on, under the goal of maximizing its contribution to video recognition. A high-capacity local CNN $f_{\textnormal{L}}$ is then adopted to process the selected patch for more accurate but computationally expensive representations. Finally, a classifier $f_{\textnormal{C}}$ integrates the features of all previous frames to produce a prediction. In the following, we describe these four components in details.

% As an overview, AdaFocus first takes a quick glance at the frames with a small light-weighted CNN, obtaining cheap but coarse global features. 

% Then these features are fed into a recurrent policy network to aggregate the temporal information and determine the location of an image patch to be focused on for each frame, under the goal of maximizing its contribution to video recognition. 

% A high-capacity local CNN will exclusively process the selected patches for more powerful but computationally expensive representations. Finally, a classifier integrates both global and local features of all frames to produce a reliable prediction.

% As shown in Figure XX, the proposed AdaFocus framework consists of four components: a light-weighted global CNN $f_{\textnormal{G}}$, a high-capacity local CNN $f_{\textnormal{L}}$, a classifier $f_{\textnormal{C}}$ and a recurrent policy network $\pi$.

\textbf{Global CNN $f_{\textnormal{G}}$ and local CNN $f_{\textnormal{L}}$}
are backbone networks that both extract deep features from the inputs, but with distinct aims. The former is designed to quickly catch a glimpse of each frame, providing necessary information for determining which region the local CNN $f_{\textnormal{L}}$ should attend to. Therefore, a light-weighted network is adopted for $f_{\textnormal{G}}$. On the contrary, $f_{\textnormal{L}}$ is leveraged to take full advantage of the selected image regions for learning discriminative representations, and hence we deploy large and accurate models. Since $f_{\textnormal{L}}$ only needs to process a series of relatively small regions instead of the full images, this stage also enjoys high efficiency. We defer the details on the architectures of $f_{\textnormal{G}}$ and $f_{\textnormal{L}}$ to Section \ref{sec:experiment}.

Formally, given video frames $\{\bm{v}_1, \bm{v}_2, \ldots\}$ with size $H\!\times\!W$, $f_{\textnormal{G}}$ directly takes them as inputs and produces the coarse global feature maps $\bm{e}^{\textnormal{G}}_{t}$: 
\begin{equation}
    \bm{e}^{\textnormal{G}}_{t} = f_{\textnormal{G}}(\bm{v}_t),\ \ \  t=1,2,\ldots,
\end{equation}
where $t$ is the frame index. By contrast, $f_{\textnormal{L}}$ processes $P\!\times\!P$ ($P<H, W$) square image patches $\{\tilde{\bm{v}}_1, \tilde{\bm{v}}_2, \ldots\}$, which are cropped from $\{\bm{v}_1, \bm{v}_2, \ldots\}$ respectively, and we have 
\begin{equation}
    \bm{e}^{\textnormal{L}}_{t} = f_{\textnormal{L}}(\tilde{\bm{v}}_t),\ \ \  t=1,2,\ldots,
\end{equation}
where $\bm{e}^{\textnormal{L}}_{t}$ denotes the fine local feature maps. Importantly, the patch $\tilde{\bm{v}}_t$ is localized to capture the most informative regions for a given task, and this procedure is fulfilled by the policy network $\pi$, which is described in the following.

\begin{figure}[t]
    % \vskip -0.1in
    \begin{center}
    \centerline{\includegraphics[width=\columnwidth]{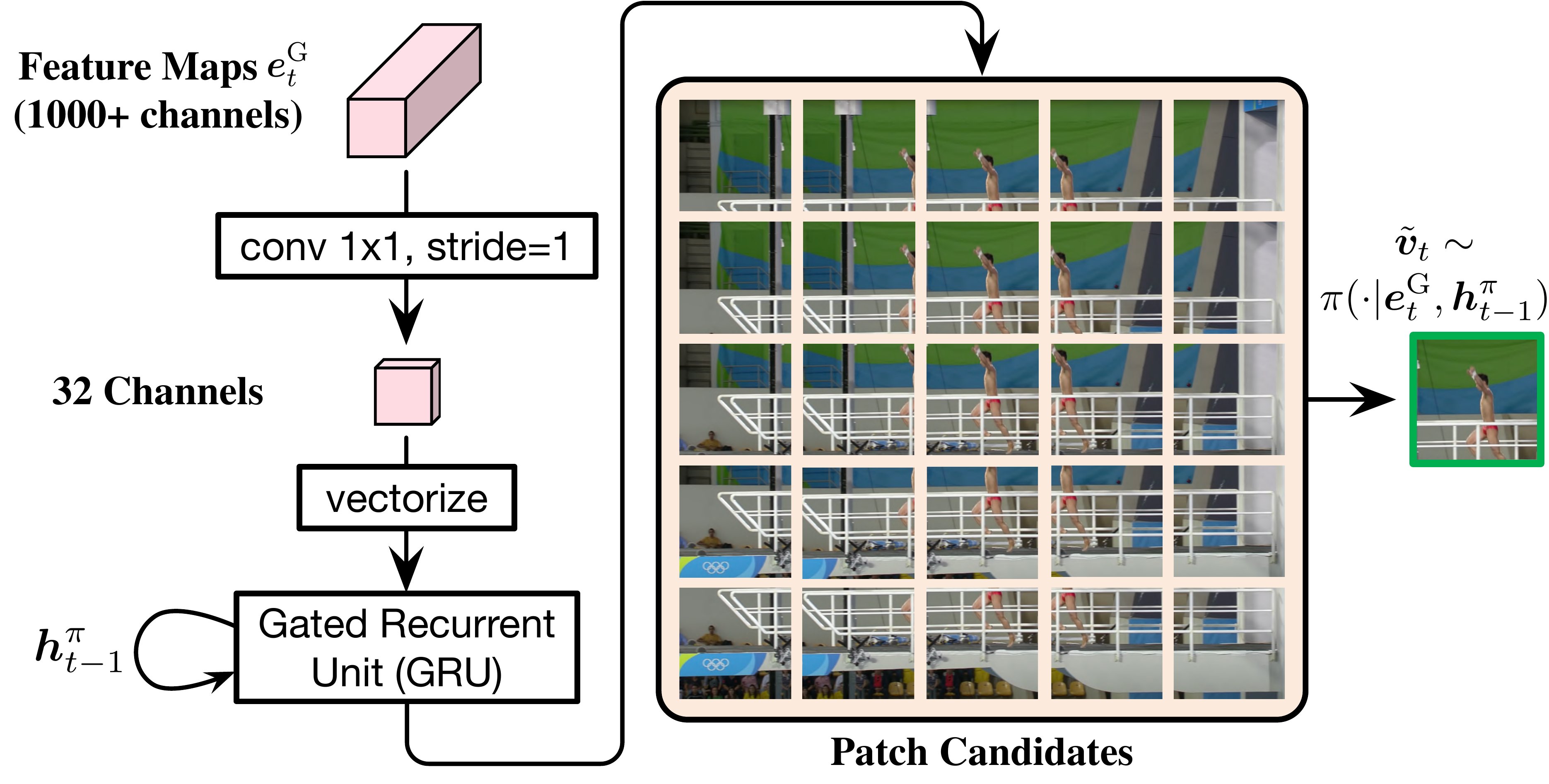}}
    % \vskip -0.025in
    \caption{\textbf{The architecture of the policy network $\pi$.} The global feature maps $\bm{e}^{\textnormal{G}}_{t}$ is fed into a 1x1 convolutional layer followed by a gated recurrent unit (GRU) to aggregate temporal information. The outputs of GRU parameterize a categorical distribution $\pi(\cdot|\bm{e}^{\textnormal{G}}_{t}, \bm{h}^{\pi}_{t-1})$ on multiple patch candidates (here we take 25 as an example). During training, we sample $\tilde{\bm{v}}_t$ from $\pi(\cdot|\bm{e}^{\textnormal{G}}_{t}, \bm{h}^{\pi}_{t-1})$, while at test time, we directly select the patch with the largest softmax probability.  \label{fig:policy_net}
    }
    \end{center}
    \vspace{-4.5ex}
    % \vskip -0.3in
\end{figure}

\textbf{Policy network $\pi$}
is a recurrent network that receives the coarse global features $\bm{e}^{\textnormal{G}}_{t}$ from the global CNN $f_{\textnormal{G}}$, and specifies which region the global CNN $f_{\textnormal{G}}$ should attend to for each frame. Note that both the information of previous and current inputs is used due to the recurrent design. Formally, $\pi$ determines the locations of images patches $\{\tilde{\bm{v}}_1, \tilde{\bm{v}}_2, \ldots\}$ to be cropped from the frames. Given that this leads to a non-differentiable operation, we formalize $\pi$ as an agent and train it with reinforcement learning. In specific, the location of the patch $\tilde{\bm{v}}_t$ is drawn from the distribution:
\begin{equation}
    \label{eq:select_location}
    \tilde{\bm{v}}_t \sim \pi(\cdot|\bm{e}^{\textnormal{G}}_{t}, \bm{h}^{\pi}_{t-1}),
\end{equation}
where $\bm{h}^{\pi}_{t-1}$ denotes the hidden states maintained in $\pi$ that are updated at $(t-1)^{\textnormal{th}}$ frame. In our implementation, we consider multiple candidates (e.g., 36 or 49) uniformly distributed across the images, and establish a categorical distribution on them, which is parameterized by the outputs of $\pi$. At test time, we simply adopt the candidate with maximum probability as $\tilde{\bm{v}}_t$ for a deterministic inference procedure. In addition, note that we do not perform any pooling on the features maps $\bm{e}^{\textnormal{G}}_{t}$ since pooling typically corrupts the useful spatial information for localizing $\tilde{\bm{v}}_t$. As an alternative, we compress the number of channels with $1\times1$ convolution to reduce the computational cost of $\pi$. An illustration of $\pi$ is shown in Figure \ref{fig:policy_net}.

\textbf{Classifier $f_{\textnormal{C}}$}
is a prediction network aiming to aggregate the information from all the frames that have been processed by the model, and output the current recognition result at each time step. To be specific, we perform global average pooling on the feature maps $\bm{e}^{\textnormal{G}}_{t}, \bm{e}^{\textnormal{L}}_{t}$ from the two aforementioned CNNs to get feature vectors $\overline{\bm{e}}^{\textnormal{G}}_{t}, \overline{\bm{e}}^{\textnormal{L}}_{t}$, and concatenate them as the inputs of $f_{\textnormal{C}}$, i.e.,
\begin{equation}
    \label{eq:classifier}
    \bm{p}_t = f_{\textnormal{C}}([\overline{\bm{e}}^{\textnormal{G}}_{1}, \overline{\bm{e}}^{\textnormal{L}}_{1}],\ldots,[\overline{\bm{e}}^{\textnormal{G}}_{t}, \overline{\bm{e}}^{\textnormal{L}}_{t}]),
\end{equation}
where $\bm{p}_t$ refers to the softmax prediction at $t^{\textnormal{th}}$ step. It is worth noting that we allow $\bm{e}^{\textnormal{G}}_{t}$ to be utilized for classification as well, with the aim of facilitating more efficient feature reusing. Such a design leverages previous observations \cite{zhou2016learning, selvaraju2017grad} revealing that CNNs are capable of achieving both remarkable localization and recognition performance at the same time. Many of existing methods also adopt similar reusing mechanisms \cite{wu2019liteeval, wu2019adaframe, meng2020ar, gao2020listen}. Besides, there are multiple possible architectures for $f_{\textnormal{C}}$. In addition to the choice of recurrent networks such as long short-term memory (LSTM) \cite{hochreiter1997long} or gated recurrent unit (GRU) \cite{cho-etal-2014-learning}, $f_{\textnormal{C}}$ can also be set as taking the average of frame-wise predictions, which are typically obtained with a common fully-connected layer, as done in \cite{lin2019tsm, meng2020ar, meng2021adafuse}.

% design choices for the architecture of $f_{\textnormal{C}}$. In addition to the 

% It can also be set as taking the average of frame-wise predictions, which are typically obtained with a common fully-connected layer, as done in \cite{lin2019tsm, meng2020ar, meng2021adafuse}. 

% Formally, given video frames $\{\bm{v}_1, \bm{v}_2, \ldots\}$, we first feed them into $f_{\textnormal{G}}$ to get the feature maps $\bm{e}^{\textnormal{G}}_{t}$:
% \begin{equation}
%     \bm{e}^{\textnormal{G}}_{t} = f_{\textnormal{G}}(\bm{v}_t), \ \ t>0,
% \end{equation}
% where $t$ is the frame index. Then we use
% Note that here we do not perform any pooling on $\bm{e}^{\textnormal{G}}_{t}$, as the spatial information it contains is important for 

\subsection{Training Algorithm}
\label{sec:training}

To ensure the four components function properly, a three-stage training algorithm is introduced. 

\textbf{Stage I: Warming-up.}
We first initialize $f_{\textnormal{G}}$, $f_{\textnormal{L}}$ and $f_{\textnormal{C}}$, but leave the policy network $\pi$ out at this stage. Then we randomly sample the image patches $\tilde{\bm{v}}_t$ to minimize the cross-entropy loss $L_{\textnormal{CE}}(\cdot)$ over the training set $\mathcal{D}_{\textnormal{train}}$:
\begin{equation}
    \label{eq:stage_1}
    \begin{split}
        \mathop{\textnormal{minimize}}_{f_{\textnormal{G}}, f_{\textnormal{L}}, f_{\textnormal{C}}}\ \ \  \mathbb{E}&_{\{\bm{v}_1, \bm{v}_2, \ldots\} \in \mathcal{D}_{\textnormal{train}}}
    \left[
        \frac{1}{T}\sum\nolimits_{t=1}^{T} L_{\textnormal{CE}}(\bm{p}_t, y)
    \right], \\ &\tilde{\bm{v}}_t\sim\textnormal{RandomCrop}(\bm{v}_t).
    % \frac{1}{\mathcal{D}_{\textnormal{train}}}\sum_{\{\bm{v}_1, \bm{v}_2, \ldots\} \in \mathcal{D}_{\textnormal{train}}}
    \end{split}
\end{equation}
Here $T$ and $y$ refer to the length and the label corresponding to the video $\{\bm{v}_1, \bm{v}_2, \ldots\}$, respectively. In this stage, the model learns to extract task-relevant information from an arbitrary sequence of frame patches, laying the basis for training the policy network $\pi$.

\textbf{Stage II: Learning to select informative patches.}
In this stage, we fix the two CNNs ($f_{\textnormal{G}}$ and $f_{\textnormal{L}}$) and the classifier $f_{\textnormal{C}}$ obtained in stage I, and evoke a randomly initialized policy network $\pi$ to be trained with reinforcement learning. Specifically, after sampling a location of $\tilde{\bm{v}}_t$ from $\pi(\cdot|\bm{e}^{\textnormal{G}}_{t}, \bm{h}^{\pi}_{t-1})$ for the frame $\bm{v}_t$ (see Eq. (\ref{eq:select_location})), $\pi$ will receive a reward $r_t$ indicating whether this action is beneficial. We train $\pi$ to maximize the sum of discounted rewards:
% The training objective of $\pi$ is to maximize the discounted sum of rewards:
\begin{equation}
    \label{eq:stage_2}
    \mathop{\textnormal{maximize}}_{\pi}\ \ \  \mathbb{E}_{\tilde{\bm{v}}_t\sim\pi(\cdot|\bm{e}^{\textnormal{G}}_{t}, \bm{h}^{\pi}_{t-1})}
    \left[
        \sum\nolimits_{t=1}^{T} \gamma^{t-1} r_t
    \right],
\end{equation}
where $\gamma\!\in\!(0, 1)$ is a discount factor for long-term rewards. In our implementation, we fix $\gamma\!=\!0.7$ and solve Eq. (\ref{eq:stage_2}) using the off-the-shelf proximal policy optimization (PPO) algorithm \cite{schulman2017proximal}. Notably, here we directly train $\pi$ on the basis of the features extracted by $f_{\textnormal{G}}$, since previous works \cite{zhou2016learning, selvaraju2017grad} have demonstrated that CNNs learned for classification generally excel at localizing task-relevant regions with their deep representations.

% , and it also performs well empirically.

Ideally, the reward $r_t$ is expected to measure the value of selecting $\tilde{\bm{v}}_t$ in terms of video recognition. With this aim, we define $r_t$ as:
\begin{equation}
    \label{eq:reward}
    \begin{split}
        &r_t(\tilde{\bm{v}}_t|\tilde{\bm{v}}_1, \ldots, \tilde{\bm{v}}_{t-1}) \\=\  &p_{ty}(\tilde{\bm{v}}_t|\tilde{\bm{v}}_1, \ldots, \tilde{\bm{v}}_{t-1}) \\  &- \ {\mathbb{E}}_{\tilde{\bm{v}}_t\sim\textnormal{RandomCrop}(\bm{v}_t)}\left[p_{ty}(\tilde{\bm{v}}_t|\tilde{\bm{v}}_1, \ldots, \tilde{\bm{v}}_{t-1})\right],
    \end{split}
\end{equation}
where $p_{ty}$ refers to the softmax prediction on $y$ (i.e., confidence on the ground truth label). When computing $r_t$, we assume all previous patches $\{\tilde{\bm{v}}_1, \ldots, \tilde{\bm{v}}_{t-1}\}$ have been determined, while only $\tilde{\bm{v}}_t$ can be changed. The second term in Eq. (\ref{eq:reward}) refers to the expected confidence achieved by the randomly sampled $\tilde{\bm{v}}_t$. By introducing it we ensures $\mathbb{E}_{\tilde{\bm{v}}_t}[r_t]=0$, which is empirically found to yield a more stable training procedure. In experiments, we estimate this term with a single time of Monte-Carlo sampling. Intuitively, Eq. (\ref{eq:reward}) encourages the model to select the patches that are capable of producing confident predictions on the correct labels with as fewer frames as possible.

\textbf{Stage III: Fine-tuning.}
At the last stage, we fine-tune $f_{\textnormal{L}}$ and $f_{\textnormal{C}}$ (or only $f_{\textnormal{C}}$) with the learned policy network $\pi$ from stage II, namely, minimizing Eq. (\ref{eq:stage_1}) with $\tilde{\bm{v}}_t\sim\pi(\cdot|\bm{e}^{\textnormal{G}}_{t}, \bm{h}^{\pi}_{t-1})$. This stage further improves the performance of our method.

% that enable the network to produce correct predictions in high conﬁdence with as fewer patches as possible.
% indicate how beneficial selecting $\tilde{\bm{v}}_t$ will be in terms of video recognition.
% which is expected to indicate how beneficial selecting $\tilde{\bm{v}}_t$ will be in terms of video recognition. 
% In other words, we always expect $\pi$

% \subsection{Compatibility with Temporal-based Methods}
% \vspace{-0.25ex}
\subsection{Reducing Temporal Redundancy}
% \vspace{-0.25ex}
\label{sec:compatibility}

% \textbf{Compatibility with temporal redundancy based techniques.}
The proposed AdaFocus processes each video frame equivalently with the same amount of computation. In fact, it is compatible with existing methods that focus on reducing the temporal redundancy of videos. To demonstrate this, we propose an extended version of AdaFocus, named as AdaFocus+, which dynamically skips less important frames for the large network $f_{\textnormal{L}}$. 

\begin{figure}[t]
    % \vskip -0.1in
    \begin{center}
    \centerline{\includegraphics[width=0.8\columnwidth]{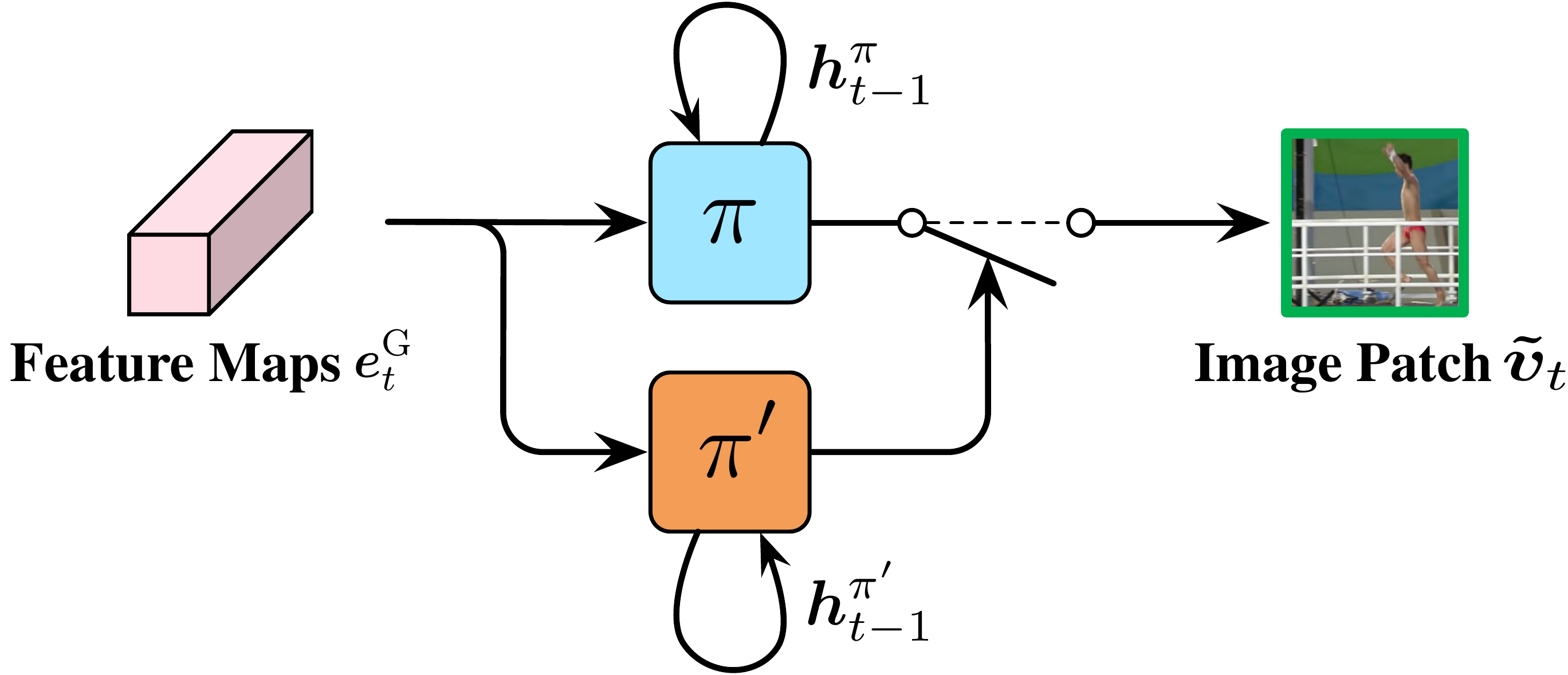}}
    % \vskip -0.025in
    \caption{\textbf{An illustration of AdaFocus+.} The proposed AdaFocus method is naturally compatible with temporal-based techniques. By involving an additional policy network $\pi'$ to control whether to attend on each frame (i.e., processing $\tilde{\bm{v}}_t$ with $f_{\textnormal{L}}$), we can further reduce the redundant computation spent on less important frames. \label{fig:temporal_agent}
    }
    \end{center}
    % \vspace{-5ex}
    \vskip -0.45in
    % \vskip -0.3in
\end{figure}

In specific, we add an additional recurrent policy network $\pi'$ that has the same inputs and architectures as $\pi$, as shown in Figure \ref{fig:temporal_agent}. This new network is trained simultaneously with $\pi$ in Stage II. For each frame, the outputs of $\pi'$ parameterize a Bernoulli random variable $b_t$:
\begin{equation}
    \label{eq:Bernoulli}
    b_t\!\sim\!\textnormal{Bernoulli}(p^{\textnormal{B}}_t), 
    \ \ \ p^{\textnormal{B}}_t\!=\!\pi'(\bm{e}^{\textnormal{G}}_{t}, \bm{h}^{\pi'}_{t-1})\!\in\!(0,1),
\end{equation}
which specifies the probability of maintaining the frame $\bm{v}_t$, i.e., $\textnormal{Pr}(b_t=1)=p^{\textnormal{B}}_t$. The hidden stated of $\pi'$, $\bm{h}^{\pi'}_{t-1}$, is updated at $(t-1)^{\textnormal{th}}$ frame.

During training, we sample $b_t$ according to Eq. (\ref{eq:Bernoulli}), and multiply it with the local feature vector $\overline{\bm{e}}^{\textnormal{L}}_{t}$, such that Eq. (\ref{eq:classifier}) changes to:
\begin{equation}
    \bm{p}_t'(b_1, \ldots, b_t) = f_{\textnormal{C}}([\overline{\bm{e}}^{\textnormal{G}}_{1}, b_1\overline{\bm{e}}^{\textnormal{L}}_{1}],\ldots,[\overline{\bm{e}}^{\textnormal{G}}_{t}, b_t\overline{\bm{e}}^{\textnormal{L}}_{t}]).
\end{equation}
In other words, if $b_t=1$, the procedure mentioned in Section \ref{sec:arch} remains unchanged. If $b_t=0$, we simply do not feed the image patch $\tilde{\bm{v}}_t$ into the local CNN $f_{\textnormal{L}}$, and concatenate $\overline{\bm{e}}^{\textnormal{G}}_{t}$ with an all-zero tensor as the inputs of the classifier $f_{\textnormal{C}}$. Notably, in this case, $\overline{\bm{e}}^{\textnormal{G}}_{t}$ will also be fed into $\pi$ to update its hidden state $\bm{h}^{\pi}_{t}$, which introduces negligible computational overhead.

\begin{figure*}[t]
    \begin{center}
    \begin{minipage}{1.37\columnwidth}
        % \centering
        \hspace{-0.05in}
        \includegraphics[width=1.015\columnwidth]{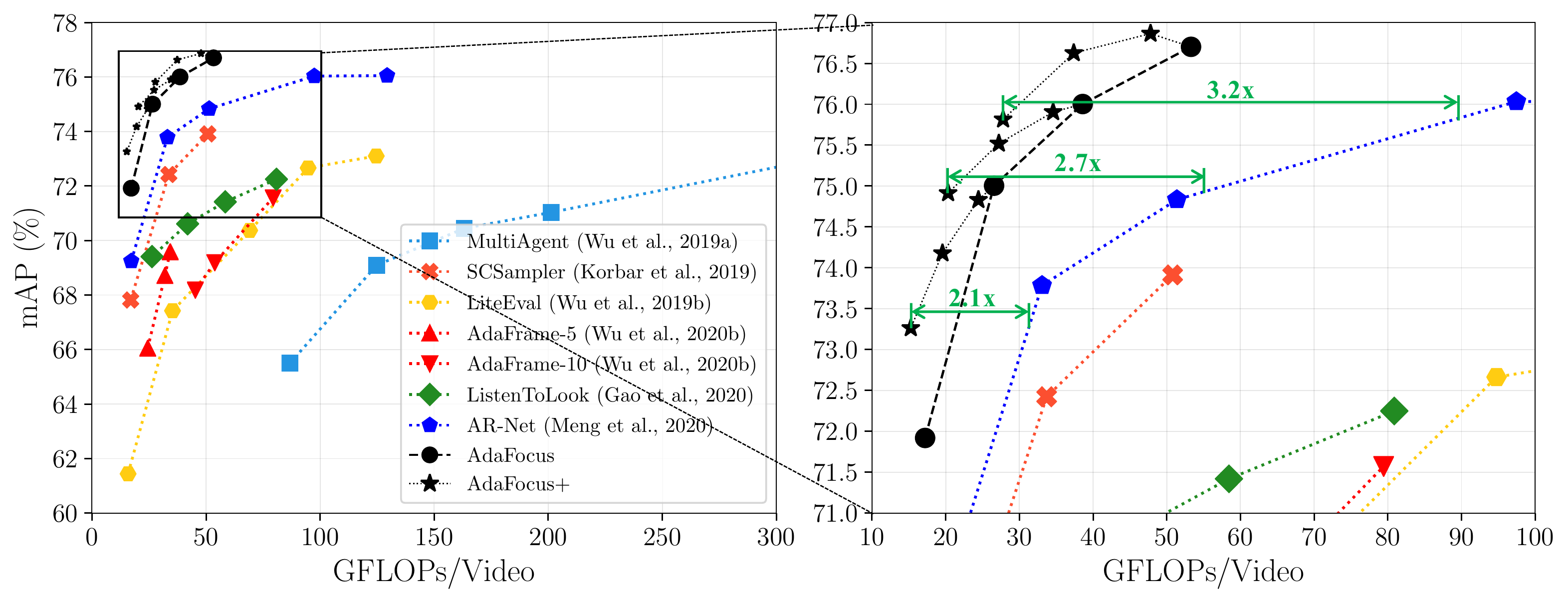}	
        \vskip -0.025in
        \caption{\textbf{Offline video recognition results on ActivityNet.} The whole video is provided at a time for a single prediction. Our method is implemented with the patch size $P^2\!\!\in$\{96$^2$, 128$^2$, 160$^2$, 192$^2$\}. AdaFocus and AdaFocus+ refer to the vanilla spatial redundancy-based AdaFocus and its augmented version by further reducing temporal redundancy. 
    }\label{fig:actnet_offline}
    % \vskip -0.2in  
    \end{minipage}
    \hspace{0.01in}
    \begin{minipage}{0.7\columnwidth}
        \vskip 0.005in
        % \centering
        \hspace{-0.1in}
        \includegraphics[width=1.0165\columnwidth]{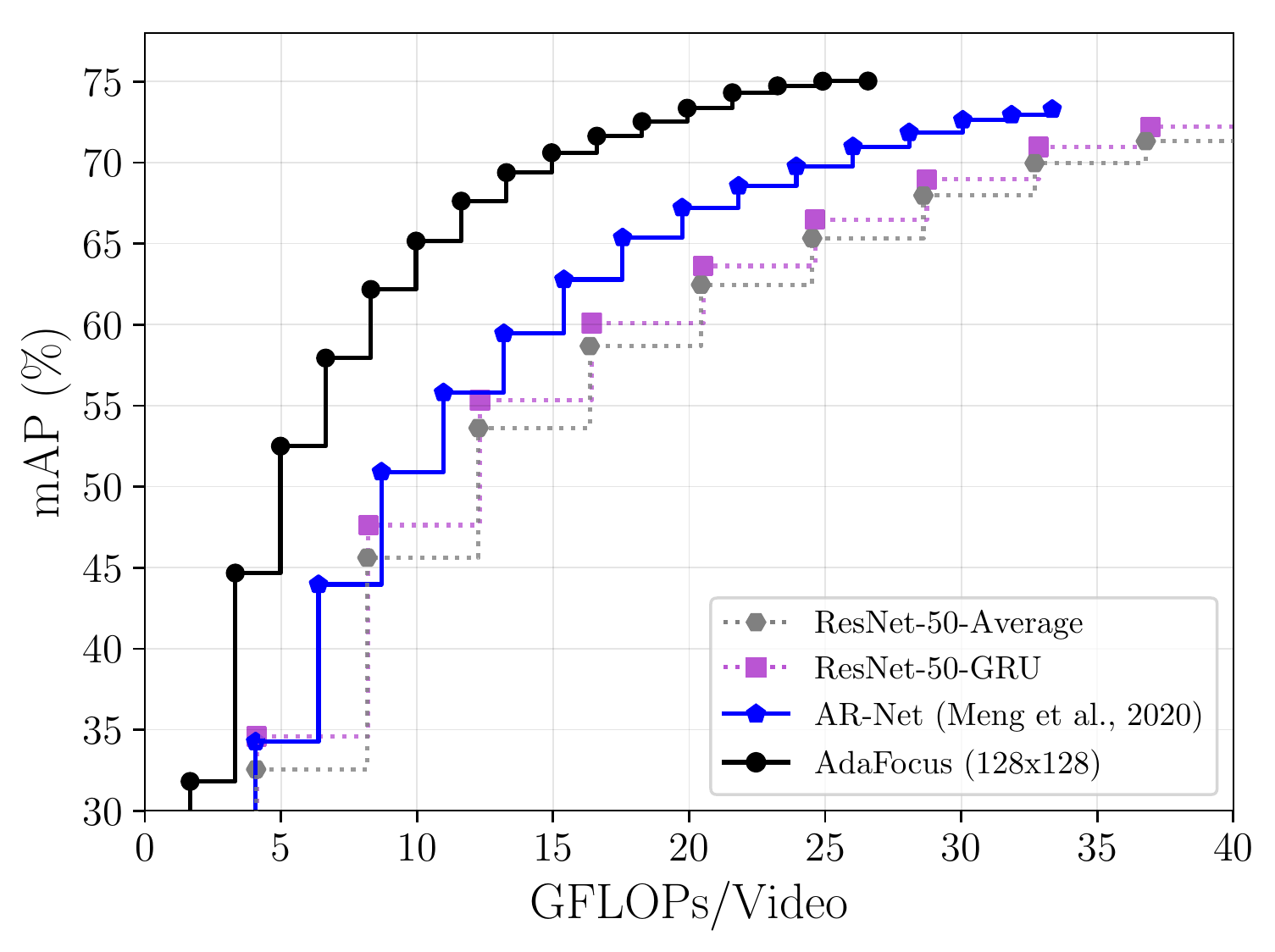}
        \vskip -0.025in
        \vskip 0.0025in	
        \caption{\textbf{Online video recognition results on ActivityNet.} Video frames come sequentially, while the models may need to output the prediction after processing any number of frames.
    }\label{fig:actnet_online}
    \end{minipage}
    \end{center}
    \vskip -0.15in
 \end{figure*}

Similarly to $\pi$, $\pi'$ is trained to maximize the sum of discounted rewards (Eq. (\ref{eq:stage_2})) as well. Here we define the reward $r_t'$ corresponding to $\pi'$ as:
% \vspace{-1.5ex}
\begin{equation}
    \begin{split}
        r_t' (b_t&|b_1, \ldots, b_{t-1}) = \\
    &\left\{
\begin{array}{lcl}
    \!\!\begin{split}
        {p}&_{ty}'(b_1, \ldots, b_{t-1}, 1)   \\
        &- {p}_{ty}'(b_1, \ldots, b_{t-1}, 0)\!-\!\lambda P^2,
    \end{split}
       & b_t=1, \vspace{0ex}\\
\!\!0,      & b_t=0,\\
\end{array} \right.
    % \begin{split}
    %     r_t' (b_t|b_1, \ldots, b_{t-1}) = &[{p}_{ty}'(b_1, \ldots, b_{t-1}, b_t) \\
    %     &- {p}_{ty}'(b_1, \ldots, b_{t-1}, 0)]\!-\!\lambda P^2,
    % \end{split}
\end{split}
\end{equation}
where ${p}_{ty}'(b_1, \ldots, b_{t-1}, 1)$ and ${p}_{ty}'(b_1, \ldots, b_{t-1}, 0)$ refer to the confidence on the ground truth label $y$ with $b_t\!\!=\!\!1$ and $b_t\!=\!0$, respectively. The coefficient $\lambda$ is a pre-defined hyper-parameter, while $P$ is the length (or width) of the patch $\tilde{\bm{v}}_t$. We use $P^2$ to estimate the required computation (FLOPs) of feeding $\tilde{\bm{v}}_t$ into $f_{\textnormal{L}}$. When $b_t\!=\!1$, the confidence gain of inferring $f_{\textnormal{L}}$ (i.e., ${p}_{ty}'(b_1, \ldots, b_{t-1}, 1)-{p}_{ty}'(b_1, \ldots, b_{t-1}, 0)$) is compared with the penalty term $\lambda P^2$, which reflects its computational costs. Only when this comparison produces positive results will the action of activating $f_{\textnormal{L}}$ be encouraged. Otherwise, $\pi'$ will be trained to decrease the probability of $b_t\!=\!1$, namely avoiding attending to any local region with the expensive $f_{\textnormal{L}}$ to avoid redundant computation.

During inference, we compare $p^{\textnormal{B}}_t$ of each frame with a fixed threshold $\rho\!\in\!(0,1)$. When $p^{\textnormal{B}}_t\!\ge\!\rho$, we process $\tilde{\bm{v}}_t$ with $f_{\textnormal{L}}$, and otherwise this patch is skipped. These two cases correspond to $b_t\!=\!1$ and $b_t\!=\!0$ during training, respectively. The value of $\rho$ should be solved on the validation set by only activating $f_{\textnormal{L}}$ for the frames with top $\eta\%$ largest $p^{\textnormal{B}}_t$ ($0\!<\!\eta\!<\!100$). One may vary $\eta\%$ for a flexible trade-off between computational costs and accuracy.

% avoiding attending to any region

% The first term $[{p}_{ty}'(b_1, \ldots, b_{t-1}, b_t)-{p}_{ty}'(b_1, \ldots, b_{t-1}, 0)]$ 
% $P$ is the side length of the square image patch $\tilde{\bm{v}}_t$, and $\lambda$ is a pre-defined coefficient. Since the 

% In addition to adaptively attend to important regions of the frames, AdaFocus-T also dynamically skips less 
% efficient video recognition methods that 
% The proposed AdaFocus framework is naturally compatible with most existing efficient video recognition methods that 
% As aforementioned, the proposed AdaFocus approach leverages spatial redundancy of video frames to improve the efficiency of video recognition. 

% \vspace{-0.25ex}
\subsection{Offline Video Recognition}
% \vspace{-0.25ex}

% \textbf{Offline video recognition.}
All the discussions above are based on the online video recognition setting where the model needs to output a reasonable prediction after seeing each frame. However, we note that AdaFocus can be straightforwardly adapted to the offline scenario where all frames are given in batch. In specific, one may train a model with the aforementioned approach, but only collect the result correspond to the last frame during inference. Importantly, the feed-forward process of both $f_{\textnormal{G}}$ and $f_{\textnormal{L}}$, which accounts for the majority of computation, can be executed in parallel, enabling an efficient implementation on GPU devices.

\section{Experiment}
% \vspace{-0.5ex}
\label{sec:experiment}

In this section, we empirically validate our method. We first compare AdaFocus with several recently proposed efficient video recognition frameworks, showing that AdaFocus gives rise to an improved efficiency. Then we implement our method by incorporating state-of-the-art light-weighted CNN architectures to demonstrate that AdaFocus complements them and further improves the efficiency. Finally, we provide detailed visualization and ablation results to give additional insights into our method. 
% Code will be available at \url{https://github.com/blackfeather-wang/AdaFocus}.

% will be released upon the acceptance of this paper.

% on five widely-used video recognition datasets, namely ActivityNet-v1.3 \cite{caba2015activitynet}, FCVID \cite{TPAMI-fcvid}, Mini-Kinetics \cite{kay2017kinetics} and Something-Something V1\&V2 \cite{goyal2017something}

\begin{table}[t]
    \centering
    \begin{footnotesize}
    \caption{\textbf{Comparisons of AdaFocus and state-of-the-art efficient video recognition frameworks on ActivityNet-v1.3 and FCVID.} GFLOPs refers to the average computational cost for processing a single video. MN2 and RN denote MobileNet-V2 and ResNet, respectively. The best results are \textbf{bold-faced}.}
    \label{tab:actnet_offline}
    % \vskip -0.03in
    \setlength{\tabcolsep}{0.7mm}{
    \vspace{5pt}
    \renewcommand\arraystretch{1}
    % \resizebox{0.95\columnwidth}{!}{
    \begin{tabular}{cccccc}
    \toprule
    \multirow{2}{*}{Methods} & \multirow{2}{*}{Backbones}  & \multicolumn{2}{c}{ActivityNet} &  \multicolumn{2}{c}{FCVID} \\
    && mAP & GFLOPs & mAP &  GFLOPs \\
    % \midrule
    \midrule
    FrameGlimpses \cite{yeung2016end} & VGG & \ \ 60.2\% & 32.9 & \ \ 71.2\% & 29.9 \\
    AdaFrame \cite{wu2019adaframe} & MN2+RN & \ \ 71.5\% & 79.0 & \ \ 80.2\% & 75.1 \\
    LiteEval \cite{wu2019liteeval} & MN2+RN & \ \ 72.7\% & 95.1 & \ \ 80.0\% & 94.3 \\
    ListenToLook \cite{gao2020listen} & MN2+RN & \ \ 72.3\%  & 81.4 & -- & -- \\
    SCSampler \cite{korbar2019scsampler} & MN2+RN & \ \ 72.9\%  & 42.0 & \ \ 81.0\% & 42.0 \\
    AR-Net \cite{meng2020ar} & MN2+RN & \ \ 73.8\% & 33.5 & \ \ 81.3\% & 35.1 \\
    \midrule
    AdaFocus (128x128)  & MN2+RN & \ \ \textbf{75.0\%} & \textbf{26.6} & \ \ \textbf{83.4\%} & \textbf{26.6} \\
    \bottomrule
    \end{tabular}}
    \end{footnotesize}
    % \vskip -0.125in
    % \vskip -0.1in
\end{table}

\textbf{Datasets.}
Our experiments are based on five widely-used video datasets:
(1) ActivityNet-v1.3 \cite{caba2015activitynet} contains 10,024 training videos and 4,926 validation videos labeled by 200 action categories. The average duration is 117 seconds;
(2) FCVID \cite{TPAMI-fcvid} includes 45,611 training videos and 45,612 validation videos labeled into 239 classes. The average duration is 167 seconds;
(3) Mini-Kinetics is a subset of Kinetics \cite{kay2017kinetics} introduced by \cite{meng2020ar, meng2021adafuse}. The dataset consists of 200 classes of videos selected from Kinetics, with 121k videos for training and 10k videos for validation;
(4) Something-Something V1\&V2 \cite{goyal2017something} are two large-scale human action datasets, including 98k and 194k videos respectively. We use the official training-validation split.

\textbf{Data pre-processing.}
Unless otherwise specified, we uniformly sample 16 frames from each video on ActivityNet, FCVID and Mini-Kinetics, while sample 8 or 12 frames on Something-Something. Following \cite{lin2019tsm, meng2020ar}, we augment training data by first adopting random scaling followed by 224x224 random cropping, and then performing random flipping on all datasets except for Something-Something V1\&V2. During inference, we resize all frames to 256x256 and centre-crop them to 224x224.

\begin{table}[t]
    \centering
    \begin{footnotesize}
    \caption{\textbf{Performance of AdaFocus and baselines on Mini-Kinetics.} GFLOPs refers to the average computational cost for processing a single video. MN2 and RN denote MobileNet-V2 and ResNet, respectively. The best results are \textbf{bold-faced}.}
    \label{tab:minik}
    % \vskip -0.03in
    \setlength{\tabcolsep}{1.5mm}{
    \vspace{5pt}
    \renewcommand\arraystretch{1}
    % \resizebox{0.95\columnwidth}{!}{
    \begin{tabular}{cccc}
    \toprule
    \multirow{2}{*}{Methods} & \multirow{2}{*}{Backbones}  & \multicolumn{2}{c}{Mini-Kinetics} \\
    && Top-1 Acc. & GFLOPs  \\
    % \midrule
    \midrule
    LiteEval \cite{wu2019liteeval} & MN2+RN & 61.0\% & 99.0 \\
    SCSampler \cite{korbar2019scsampler} & MN2+RN & 70.8\% & 42.0  \\
    AR-Net \cite{meng2020ar} & MN2+RN & 71.7\% & 32.0  \\
    \midrule
    AdaFocus (128x128)  & MN2+RN & {72.2\%} & {26.6}  \\
    AdaFocus (160x160)  & MN2+RN & \textbf{72.9\%} & {38.6}  \\
    AdaFocus+ (160x160)  & MN2+RN & {71.7\%} & \textbf{20.3}  \\
    \bottomrule
    \end{tabular}}
    \end{footnotesize}
    % \vskip -0.125in
    % \vskip -0.1in
\end{table}

\begin{table*}[!t]
    \centering
    \begin{footnotesize}
    \caption{\textbf{Performance of AdaFocus-TSM and other recently proposed efficient CNNs on Something-Something (Sth-Sth)}. TSM+ refers to the augmented TSM baseline with the same network architecture as our method except for the policy network $\pi$. We uniformly sample 8/12 frames for the MobileNet/ResNet-50 in our models\textcolor{red}{\protect\footnotemark[2]}. The latency and throughput are tested on a 2.20GHz Intel Core i7-10870H CPU and a NVIDIA GeForce RTX 2080Ti GPU with the batch size of 1 and 64, respectively. The best results are \textbf{bold-faced}.}
    \label{tab:sthsth}
    % \vskip -0.04in
    \setlength{\tabcolsep}{0mm}{
    \vspace{5pt}
    \renewcommand\arraystretch{1}
    \resizebox{2.07\columnwidth}{!}{
    \begin{tabular}{ccccccccc} 
    \toprule
    \multirow{2}{*}{{Method}} & \multirow{2}{*}{{Backbones}}  & \multirow{2}{*}{{\#Frames}}  & \multicolumn{2}{c}{{Sth-Sth V1}}  & \multicolumn{2}{c}{{Sth-Sth V2}} & Latency & Throughput\\ %\multicolumn{2}{c}{Practical Efficiency}\\
    % &&&&&&&& \\
    % \vspace{-0.1em}
    &&& \footnotesize{{\ \ Top-1 Acc.}} & \footnotesize{{GFLOPs}} &  \footnotesize{{Top-1 Acc.}} & \footnotesize{{GFLOPs}}  &   \scriptsize{(Intel Core i7, bs=1)} & \scriptsize{(NVIDIA 2080Ti, bs=64)}\\  
    % \midrule
    \midrule
    I3D \cite{carreira2017quo} & 3DResNet50 & 32$\times$2  & 41.6\% &306 &  - & - &  - & -\\
    I3D+GCN+NL \cite{wang2018videos} & 3DResNet50 & 32$\times$2 & 46.1\% & 606 & - & - &  - & -\\
    ECO\textsubscript{En}Lite \cite{zolfaghari2018eco} & BN-Inception + 3DResNet18 & 92  & {46.4\%} & 267 & - & - &  - & - \\
    \midrule
    % TSN \cite{wang2016temporal} & BN-Inception & 8& 10.7M & 19.5 & 16.0G & 16.0G & 33.4 &  \\ 
    TSN \cite{wang2016temporal} & ResNet50 & 8 & 19.7\% & 33.2 &27.8\% &  33.2  &  - & - \\ 
    % TRN\textsubscript{Multiscale} \cite{zhou2018temporal} & BN-Inception & 8 & 18.3M& 34.4 & 16.0G & 16.0G& 48.8 & \\
    TRN\textsubscript{RGB/Flow} \cite{zhou2018temporal} & BN-Inception & 8/8  & 42.0\% & 32.0 & 55.5\% & 32.0   &  - & - \\
    ECO \cite{zolfaghari2018eco} & BN-Inception+3DResNet18 & 8 & 39.6\% & 32.0 & - & - &  - & -\\
    AdaFuse \cite{meng2021adafuse} & ResNet50 & 8 & 46.8\% & 31.5 & 59.8\% & 31.3  &  - & - \\
    TSM \cite{lin2019tsm}  & ResNet50 & 8 & 46.1\% & 32.7 &  59.1\% & 32.7  & 0.32s &  \ \ 128.8 Videos/s\ \ \\
    \midrule
    TSM+ \cite{lin2019tsm}  & MobileNet-V2+ResNet50 & \ \ 8+8\textcolor{red}{\protect\footnotemark[2]}  & 47.0\% & 35.1 & 59.6\% & 35.1 & 0.42s &  \ \ 105.0 Videos/s\ \    \\
    AdaFocus-TSM (144x144)  & MobileNet-V2+ResNet50 & 8+12 & 47.0\% & \textbf{23.5} (\textcolor{blue}{$\downarrow$1.49x}) & 59.7\%  & \textbf{23.5} (\textcolor{blue}{$\downarrow$1.49x})  & \ \ \ \ \ \textbf{0.32s} (\textcolor{blue}{$\downarrow$1.31x}) &  \ \ \ \ \textbf{143.8 Videos/s} (\textcolor{blue}{$\uparrow$1.37x})   \\
    AdaFocus-TSM (160x160)  & MobileNet-V2+ResNet50 & 8+12 & 47.6\% & 27.5 & 60.2\% &  27.5 & 0.36s &  \ \ 122.1 Videos/s\ \    \\
    AdaFocus-TSM (176x176)  & MobileNet-V2+ResNet50 & 8+12 & \textbf{48.1\%} & 33.7 & \textbf{60.7\%} & 33.7  & 0.42s &  \ \ 104.2 Videos/s\ \    \\
    \bottomrule
    \end{tabular}}}
    \end{footnotesize}
    % \vskip -0.05in
\end{table*}

% \vspace{-0.75ex}
\subsection{Comparisons with State-of-the-art Efficient Video Recognition Methods}
% \vspace{-0.75ex}
\textbf{Baselines.}
In this subsection, AdaFocus is compared with several competitive baselines that focus on facilitating efficient video recognition, including MultiAgent \cite{wu2019multi}, SCSampler \cite{korbar2019scsampler}, LiteEval \cite{wu2019liteeval}, AdaFrame \cite{wu2019adaframe}, Listen-to-look \cite{gao2020listen} and AR-Net \cite{meng2020ar}. Due to spatial limitations, we briefly introduce them in Appendix A.

% (1) MultiAgent \cite{wu2019multi} proposes to learn to select important frames with Multi-agent reinforcement learning.
% (2) SCSampler \cite{korbar2019scsampler} introduces a light-weighted framework to efficiently identify the most salient temporal clips within a long video. We follow the implementation of \cite{meng2020ar}.
% (3) LiteEval \cite{wu2019liteeval} combines a coarse LSTM and a fine LSTM to adaptively allocate computation based on the importance of frames.
% (4) AdaFrame \cite{wu2019adaframe} learns to dynamically select informative frames with reinforcement learning and performs adaptive inference.
% (5) Listen-to-look \cite{gao2020listen} fuses image and audio information to select the key clips within a video. As we do not leverage the audio of videos, for a fair comparison, we adopt its image-based version introduced in their paper.
% (6) AR-Net \cite{meng2020ar} dynamically identifies the importance of video frames, and process them with different resolutions accordingly.

\textbf{Implementation details.}
We deploy MobileNet-V2 \cite{sandler2018mobilenetv2} and ResNet-50 \cite{he2016deep} as the global CNN $f_{\textnormal{G}}$ and local CNN $f_{\textnormal{L}}$ in AdaFocus. A one-layer gated recurrent unit (GRU) \cite{cho-etal-2014-learning} with a hidden size of 1024 is used in both the policy network $\pi$ and the classifier $f_{\textnormal{C}}$. The number of patch candidates is set to 49 (uniformly distributed in 7x7). Due to the limited space, training details are deferred to Appendix B.

% These two types of networks are also exploited as backbones in most of the baselines \cite{korbar2019scsampler, wu2019liteeval, wu2019adaframe, gao2020listen, meng2020ar}. 

\textbf{Offline video recognition.}
We first implement AdaFocus under the offline recognition setting, where our method produces a single prediction after processing the whole video. This setting is adopted by the papers of most baselines as well. The results on ActivityNet and FCVID are presented in Table \ref{tab:actnet_offline}. We use a patch size of $P^2\!=\!128^2$ in AdaFocus, and evaluate the performance of different methods via mean average precision (mAP) following the common practice \cite{wu2019adaframe, wu2019liteeval, gao2020listen, meng2020ar} on these two datasets. It can be observed that our method outperforms the alternative baselines by large margins in terms of efficiency. For example, on FCVID, AdaFocus achieves 2.1\% higher mAP (83.4\% v.s.\! 81.3\%) than the strongest baseline, AR-Net, with 1.3x less computation (26.6 GFLOPs v.s.\! 33.5 GFLOPs).

\textbf{Results of varying patch sizes}
are presented in Figure \ref{fig:actnet_offline}. We change the patch size within $P^2\in$\{96x96, 128x128, 160x160, 192x192\}, and plot the corresponding mAP v.s. FLOPs relationships in black dots. We also present the variants of baselines with various computational costs. One can observe that AdaFocus leads to a considerably better trade-off between efficiency and accuracy.

\begin{figure*}[t]
    % \vskip -0.1in
    \begin{center}
    \centerline{\includegraphics[width=2.1\columnwidth]{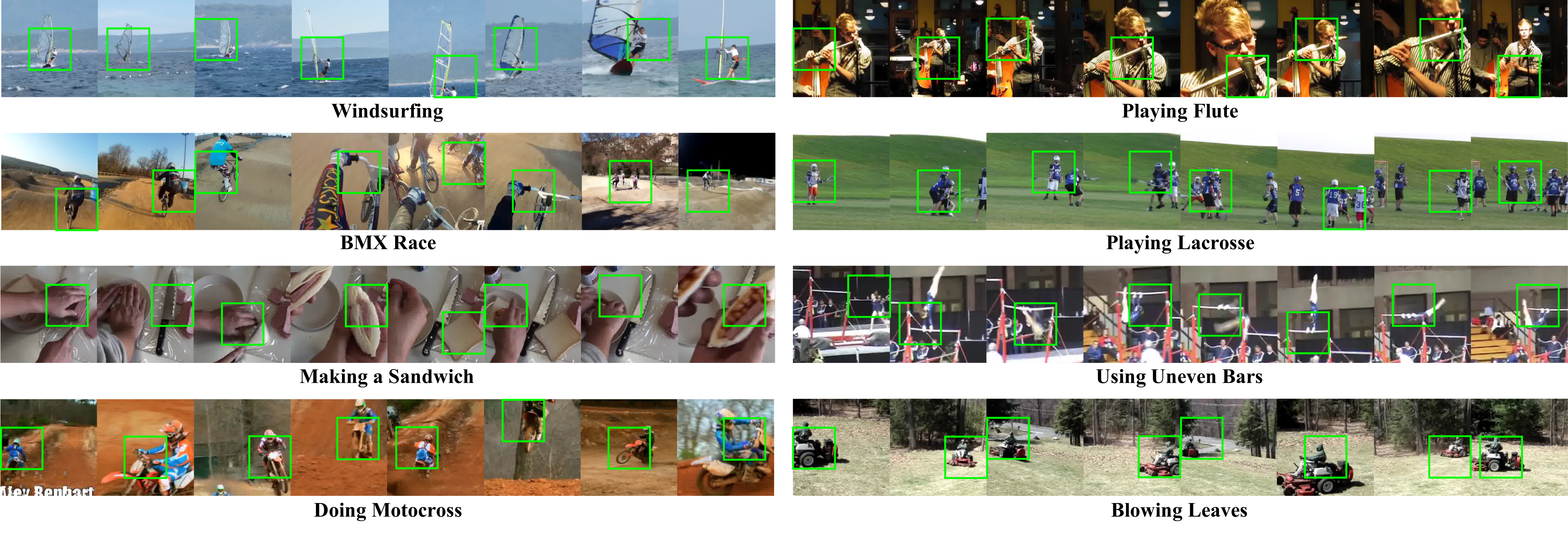}}
    % \vskip -0.06in
    \caption{\textbf{Visualization results (zoom in for details).} The green boxes indicate the locations of the image patches selected by AdaFocus. \label{fig:Visualization}
    }
    \end{center}
    \vspace{-4ex}
    % \vskip -0.3in
\end{figure*}

\footnotetext[2]{In fact, we can also sample 8/12 frames for TSM+, but this increases computational costs dramatically \!($\sim\!\!1.5$x). Hence, we do not consider it.}

\textbf{Improvements from further reducing temporal redundancy.}
Then we test extending AdaFocus by skipping less informative frames, as stated in Section \ref{sec:compatibility}. The results are presented as AdaFocus+ (black stars) in Figure \ref{fig:actnet_offline}. The coefficient $\lambda$ is set to $1e-6$, while the skipping proportion $\eta\%$ is varied within \{0.9, 0.7, 0.5\}. For the ease of implementation, we solve the threshold $\rho$ on the training set. We find this achieves almost the same performance as using a validation set. It is clear that further reducing temporal redundancy leads to a significantly better efficiency. With a given mAP, the number of required GFLOPs per video for AdaFocus+ is approximately 2.1-3.2x less than AR-Net.

\textbf{Results on Mini-Kinetics}
are presented in Table~\ref{tab:minik}. The observations here are similar to ActivityNet/FCVID. AdaFocus+ reduces the required computation to reach 71.7\% accuracy by 1.6x (20.3 GFLOPs v.s. 32.0 GFLOPs).

% outperforms the competitive baselines by more than $0.5\%$ in Top-1 accuracy with at least 1.2x less computation (26.6 GFLOPs v.s. 32.0 GFLOPs).

\textbf{Online video recognition}
results are shown in Figure \ref{fig:actnet_online}. Note that here we assume a stream of video frames come in sequentially for processing and the model may need to output a prediction at any time. In specific, we take a fixed number of frames from the beginning of videos, feed them into networks to evaluate the results, and change the number of frames to obtain the mAP-FLOPs trade-off. We consider two additional baselines: (1) ResNet-50-Average averages the frame-level predictions of a ResNet-50 with the full inputs; (2) ResNet-50-GRU augments (1) by aggregating the features across frames using a GRU classifier. Figure \ref{fig:actnet_online} shows that our method is able to obtain much better performance given the same number of FLOPs, which enables accurate and fast recognition in real-time applications.

% feed a fixed number of beginning frames into the networks, and change this the number of frames to obtain a mAP-FLOPs curve. 
% a prediction may be required after processing any number of frames.
% AdaFocus+ save 
% reduces the number of required Multiply-Adds for the given test accuracy by approximately 2 − 3× times.
% the strongest baseline, AR-Net achieves a mAP of 73.8\% at the cost of 33.5 GFLOPs per video, while AdaFocus outperforms
% which is adopted by most of the baselines. 

\begin{table}[t]
    \centering
    \begin{footnotesize}
    \caption{\textbf{Effects of Reusing $\bm{e}^{\textnormal{G}}_{t}$ for Recognition.}}
    \label{tab:abl_1}
    % \vskip -0.04in
    \setlength{\tabcolsep}{1.5mm}{
    \vspace{5pt}
    \renewcommand\arraystretch{1}
    % \resizebox{0.78\columnwidth}{!}{
    \begin{tabular}{c|cccc}
    \toprule
    Reusing $\bm{e}^{\textnormal{G}}_{t}$ &  \multicolumn{4}{|c}{mAP} \\
    for Recognition & 96x96 & 128x128 & 160x160 & 192x192 \\
    % \midrule
    \midrule
    \xmark & {70.2\%} & {73.4\%} & {75.0\%} & {75.9\%} \\
    \cmark & \ \textbf{71.9\%} & \ \textbf{75.0\%} & \ \textbf{76.0\%} & \ \textbf{76.7\%} \\

    % -&-&-&& \\
    % \midrule
    % -&-&\checkmark& 65.76 & 70.69 \\
    % -&\checkmark&\checkmark& 70.19 & 73.43 \\
    % \checkmark&\checkmark&\checkmark&\textbf{71.92}  & \textbf{75.00}  \\
    \bottomrule
    \end{tabular}}
    \end{footnotesize}
    % \vskip -0.1in
    % \vskip -0.21in
\end{table}

% \vspace{-0.75ex}
\subsection{Building on Top of Efficient CNNs}
% \vspace{-0.75ex}

 \textbf{Setup.}
 In this subsection, we implement AdaFocus on top of the recently proposed efficient network architecture, CNNs with temporal shift module (TSM) \cite{lin2019tsm}, to demonstrate that our method can effectively improve the efficiency of such state-of-the-art light-weighted models. Specifically, we still use MobileNet-V2 and ResNet-50 as $f_{\textnormal{G}}$ and $f_{\textnormal{L}}$, but add TSM to them. A fully-connected layer is deployed as the classifier $f_{\textnormal{C}}$, and we average the frame-wise predictions as the output, following the design of TSM \cite{lin2019tsm}. For a fair comparison, we augment the vanilla TSM by introducing the same two backbone networks as ours (named as TSM+), where their output features are also concatenated to be fed into a linear classifier. In other words, TSM+ differentiates itself from AdaFocus only in that it feeds the whole frames into ResNet-50, while we feed the selected image patches.

 The setting of offline video recognition on Something-Something V1\&V2 is used here. Notably, as the videos in the two datasets are very short (average duration $\!\approx\!$ 4s), we find the networks require the visual angle of adjacent inputs (frames/patches) to be similar for high generalization performance. We also observe that the locations of task-relevant regions do not significantly change across the frames in the same video. Therefore, here we let AdaFocus generate a single patch location for the whole video after aggregates the information of all frames. Importantly, such a simplification does not affect the main idea of our method since different videos have varying patch locations. The architecture of the policy network $\pi$ and the training algorithm remain unchanged. Details of training hyper-parameters are deferred to Appendix B.

%  All other network architectures and training algorithms remain unchanged. More details are deferred to Appendix X due to the limited space.
%  For a fair comparison, these two networks are also deployed

%  we also deploy these two networks in the basic TSM, and concatenate the output features for recognition, named as TSM+. 

%  The setting of offline video recognition on Something-Something V1\&V2 is considered here. Notably, we find that as the two datasets consist of very short videos (average duration $\approx$ 4s), the networks require the visual angle of adjacent inputs (frames/patches) to be similar for high generalization performance. Therefore, here we let AdaFocus generate a single patch location for the whole video after aggregates the information of all frames. Importantly, such a simplification does not affect the main idea of AdaFocus since different videos have varying patch locations, while this already performs well empirically. 

%  Besides, we deploy a fully-connected layer as  the classifier $f_{\textnormal{C}}$, and average the frame-wise predictions as the output, following the vanilla TSM \cite{lin2019tsm}. 
 
% All other network architectures and training algorithms remain unchanged. More details are deferred to Appendix X due to the limited space.
\textbf{Results on Something-Something}
 are reported in Table \ref{tab:sthsth}. One can observe that by reducing the input size of the relatively expensive ResNet-50 network, AdaFocus enables TSM to process more frames in the task-relevant region of each video using the same computation, leading to a significantly improved efficiency. For example, AdaFocus achieves the same performance as TSM+ with 1.5x less GFLOPs on Something-Something V1.
 
 \textbf{Practical efficiency.}
 In Table \ref{tab:sthsth}, we also test the actual inference speed of AdaFocus-TSM on both an Intel i7 CPU and a NVIDIA 2080Ti GPU, with the batch size of 1 and 64, respectively, which are sufficient to saturate the two devices. It can be observed that our practical speedup is significant as well, with a slight drop compared with theoretical results. We tentatively attribute this to the inadequate hardware-oriented optimization in our implementation.
 
%  We also present the Top-1 accuracy and FLOPs of several other efficient CNN as baselines. 
 
%  For a fair comparison, we deploy both MobileNet-V2 and ResNet in the basic TSM, and concatenate the output features for recognition, named as TSM+. 

%  One can observe from Table \ref{tab:sthsth} that by reducing the input size of the relatively expensive ResNet-50 network, AdaFocus enables TSM to process more frames in the task-relevant region of each video using the same computation, leading to a significantly improved efficiency. For example, AdaFocus-TSM achieves the same performance as TSM+ with 1.5x less FLOPs on Something-Something V1.
 
% \vspace{-0.5ex}
\subsection{Analytical Results}
% \vspace{-0.5ex}

\textbf{Visualization.}
In Figure~\ref{fig:Visualization}, we visualize the regions selected by our proposed AdaFocus. Here we uniformly sample 8 video frames from ActivityNet. One can observe that our method effectively guides the expensive local CNN $f_{\textnormal{L}}$ to attend to the task-relevant regions of each frame, such as the sailboard, bicycle and flute.

\textbf{Importance of reusing $\bm{e}^{\textnormal{G}}_{t}$ for recognition.}
As aforementioned, our method effectively leverages the coarse global feature $\bm{e}^{\textnormal{G}}_{t}$ for both localizing the task-relevant patch $\tilde{\bm{v}}_t$ and recognition. As shown in Table~\ref{tab:abl_1}, only using $\bm{e}^{\textnormal{G}}_{t}$ for localization degrades the mAP by $1-1.5\%$, which demonstrates the effects of this reusing mechanism.

\begin{table}[t]
    \centering
    \begin{footnotesize}
    % \vskip -0.05in
    \caption{\textbf{Comparisons of various patch selection policies.} Fixed policies are pre-defined without leveraging reinforcement learning (RL). For RL-based policy, we change the design of rewards.}
    % \vskip -0.04in
    \label{tab:abl_2}
    \setlength{\tabcolsep}{0.85mm}{
    \vspace{5pt}
    \renewcommand\arraystretch{1}
    % \resizebox{0.9\columnwidth}{!}{
    \begin{tabular}{c|c|cccc}
    \toprule
    \multicolumn{2}{c}{\multirow{2}{*}{Ablation}}&  \multicolumn{4}{|c}{mAP} \\
    \multicolumn{2}{c|}{\ \ }  & \ 96x96\  & \ 128x128\  & \ 160x160\  & \ 192x192\  \\
    % \midrule
    \midrule
    \multirow{3}{*}{\shortstack{Fixed\\Policy}}& Random Policy & \ 65.8\%\  & \ 70.7\%\  & \ 73.1\%\  & \ 74.8\%\  \\
    & Central Policy & \ 61.9\%\  & \ 68.7\%\  & \ 72.4\%\  & \ 74.8\%\  \\
    & Gaussian Policy & \ 64.7\%\  & \ 70.6\%\  & \ 73.5\%\  & \ 74.9\%\  \\
    \midrule
    \multirow{3}{*}{\shortstack{Learned\\Policy\\by RL}}& Confidence Reward & \ 68.5\%\  & \ 72.3\%\  & \ 74.1\%\  & \ 75.5\%\  \\
    & Increments Reward & \ 69.4\%\  & \ 72.7\%\  & \ 74.4\%\  & \ 75.6\%\  \\
    & AdaFocus (ours) & \ \ \textbf{70.2\%}\  & \ \ \textbf{73.4\%}\  & \ \ \textbf{75.0\%}\  & \ \ \textbf{75.9\%}\  \\

    % -&-&-&& \\
    % \midrule
    % -&-&\checkmark& 65.76 & 70.69 \\
    % -&\checkmark&\checkmark& 70.19 & 73.43 \\
    % \checkmark&\checkmark&\checkmark&\textbf{71.92}  & \textbf{75.00}  \\
    \bottomrule
    \end{tabular}}
    \end{footnotesize}
    % \vskip -0.125in
    % \vskip -0.2in
\end{table}

\textbf{Effectiveness of the learned patch selection policy}
is validated in Table~\ref{tab:abl_2}. We consider its three alternatives: (1) \emph{randomly} sampling patches, (2) cropping patches from the \emph{centres} of the frames, and (3) sampling patches from a standard \emph{gaussian} distribution centred the frame. In addition, we test altering the reward function for reinforcement learning to: (1) \emph{confidence reward} directly uses the confidence on ground truth labels as rewards, and (2) \emph{increments reward} utilizes the increments of confidence as rewards. For a clear comparison, here we do not reuse $\bm{e}^{\textnormal{G}}_{t}$ for recognition. An interesting phenomenon is that random policy appears strong and outperforms the central policy, which may be attributed to the spatial similarity between frames. That is to say, adjacent central patches might have repetitive contents, while randomly sampling is likely to collect more comprehensive information. Besides, it is shown that the learned policies have considerably better performance, and our proposed reward function significantly outperforms others.

%  leverage more temporal informaion

\vspace{-0.75ex}
\section{Conclusion}
\vspace{-0.75ex}

This paper has proposed a \emph{spatial redundancy} based approach for efficient video recognition, AdaFocus. Inspired by the fact that not all image regions in video frames are task-relevant, AdaFocus reduces computational costs by inferring the high-capacity network only on a small but informative patch of each frame, which is adaptively localized with reinforcement learning. We further show that our method can be extended by dynamically skipping less valuable frames. Extensive experiments demonstrate that our method outperforms existing works in terms of both theoretical computational efficiency and actual inference speed.

\section*{Acknowledgements}
This work is supported in part by the National Science and Technology Major Project of the Ministry of Science and Technology of China under Grants 2018AAA0100701, the National Natural Science Foundation of China under Grants 61906106 and 62022048, the Institute for Guo Qiang of Tsinghua University and Beijing Academy of Artificial Intelligence.

{\small
\bibliographystyle{ieee_fullname}
\bibliography{egbib}
}

% \newpage
% \newpage
\appendix

%%%%%%%%% BODY TEXT
\section*{Appendix for ``Adaptive Focus for Efficient Video Recognition''}
\section{Introduction of Baselines}
AdaFocus is compared with several competitive baselines that focus on facilitating efficient video recognition, including MultiAgent \cite{wu2019multi}, SCSampler \cite{korbar2019scsampler}, LiteEval \cite{wu2019liteeval}, AdaFrame \cite{wu2019adaframe}, Listen-to-look \cite{gao2020listen} and AR-Net \cite{meng2020ar}.
\begin{itemize}
   \item MultiAgent \cite{wu2019multi} proposes to learn to select important frames with multi-agent reinforcement learning.
   \item SCSampler \cite{korbar2019scsampler} introduces a light-weighted framework to efficiently identify the most salient temporal clips within a long video. We follow the implementation of \cite{meng2020ar}.
   \item LiteEval \cite{wu2019liteeval} combines a coarse LSTM and a fine LSTM to adaptively allocate computation based on the importance of frames.
   \item AdaFrame \cite{wu2019adaframe} learns to dynamically select informative frames with reinforcement learning and performs adaptive inference.
   \item Listen-to-look \cite{gao2020listen} fuses image and audio information to select the key clips within a video. As we do not leverage the audio of videos, for a fair comparison, we adopt its image-based version introduced in their paper.
   \item AR-Net \cite{meng2020ar} dynamically identifies the importance of video frames, and processes them with different resolutions accordingly.
\end{itemize}

\section{Implementation Details}

\subsection{Training Hyper-parameters for Section 4.1}

In our implementation, we always train $f_{\textnormal{G}}$, $f_{\textnormal{L}}$ and $f_{\textnormal{C}}$ using a SGD optimizer with cosine learning rate annealing and a Nesterov momentum of 0.9 \cite{He_2016_CVPR, huang2019convolutional, huang2017snapshot, wang2019implicit, wang2021revisiting, lin2019tsm, meng2020ar}. The size of the mini-batch is set to 64, while the L2 regularization coefficient is set to 1e-4. We initialize $f_{\textnormal{G}}$ and $f_{\textnormal{L}}$ by fine-tuning the ImageNet pre-trained  MobileNet-V2 \cite{sandler2018mobilenetv2} and ResNet-50 \cite{he2016deep} (we use the official models provided by PyTorch) using full inputs for 15 epochs with an initial learning rate of 0.01. In stage I, we train $f_{\textnormal{L}}$ and $f_{\textnormal{C}}$ using randomly sampled patches for 50 epochs with an initial learning rate of 5e-4 and 0.05, respectively. Here we do not train $f_{\textnormal{G}}$ as we find this does not significantly improve the performance, but increases the training time. In stage II, we train $\pi$/$\pi'$ with an Adam optimizer \cite{kingma2014adam} for 50/10 epochs. The same training hyper-parameters as \cite{wang2020glance} are adopted. In stage III, we only fine-tune $f_{\textnormal{C}}$ with the learned policy for 10 epochs, since we find further fine-tuning $f_{\textnormal{L}}$ leads to trivial improvements but prolongs the training time. The initial learning rates are set to 5e-4 and 5e-3 for Mini-Kinetics and ActivityNet/FCVID, respectively.

\subsection{Training Hyper-parameters for Section 4.2}

Here we initialize $f_{\textnormal{G}}$ and $f_{\textnormal{L}}$ by training them using the same configuration as \cite{lin2019tsm}. The training procedure of AdaFocus is the same as Section 4.1 except for the following changes. In stage I, we use the initial learning rate of 1e-5 and 0.01 for $f_{\textnormal{L}}$ and $f_{\textnormal{C}}$, respectively, and train them for 10 epochs. In stage III, we use an initial learning rate of 5e-4 for $f_{\textnormal{C}}$. Note that TSM+ follows exactly the same training procedure as our method. The only difference is that TSM+ does not train the policy network $\pi$, since it adopts full frames as inputs.

\end{document}